\documentclass{article}

\usepackage[preprint]{nips_2018}

\usepackage[utf8]{inputenc} 
\usepackage[T1]{fontenc}    
\usepackage{hyperref}       
\usepackage{url}            
\usepackage{booktabs}       
\usepackage{amsfonts}       
\usepackage{nicefrac}       
\usepackage{microtype}      

\usepackage{times}
\usepackage{latexsym}
\usepackage{float}
\usepackage{xcolor}
\usepackage{graphicx}
\usepackage{dsfont}
\usepackage{amsmath}
\usepackage{flushend}
\usepackage{caption}
\usepackage{natbib}
\usepackage{url}

\title{Leveraging User Engagement Signals For Entity Labeling in a Virtual Assistant}

\author{
  Deepak Muralidharan\thanks{\textbf{Equal contributions. Alphabetically ordered.}}, \enspace Justine Kao\footnote[1]{},\enspace Xiao Yang\footnote[1]{},\enspace  Lin Li, \\  \textbf{Lavanya  Viswanathan,\enspace   Mubarak Seyed Ibrahim,\enspace  Kevin Luikens,} \\ \textbf{Stephen Pulman,\enspace  Ashish Garg,\enspace Atish Kothari, \enspace Jason Williams} \\
  Apple Inc., One Apple Park Way, Cupertino, CA 95014}

\begin{document}

\maketitle

\begin{abstract}

Personal assistant AI systems such as Siri, Cortana, and Alexa have become widely used as a means to accomplish tasks through natural language commands. 
 However, components in these systems generally rely on supervised machine learning algorithms that require large amounts of hand-annotated training data, which is expensive and time-consuming to
  collect.
The ability to incorporate unsupervised, weakly supervised, or distantly supervised data holds
  significant promise in overcoming this bottleneck.
  In this paper, we describe a framework that leverages user engagement signals (user behaviors that demonstrate a positive or negative response to content) to automatically create granular entity labels for training data augmentation.
Strategies such as multi-task learning and validation using an external knowledge base are employed to incorporate the engagement-annotated data and to boost the model's accuracy on a sequence labeling task. 
Our results show that learning 
  from data automatically labeled by user engagement signals achieves
  significant accuracy gains in a production deep learning system, when measured on both the sequence labeling task as well as on user-facing results produced by the system end-to-end.
We believe this is the first use of user engagement signals to help generate training data for a sequence labeling task on a large scale, and can be applied in practical settings to speed up new feature deployment when little human-annotated data is available.

\end{abstract}

\section{Introduction}
Accomplishing a voice controlled task using
a virtual assistant agent such as Siri, Cortana, or Alexa usually involves several steps. First, a speech recognition module converts audio signals into text. Next, a natural language understanding (NLU) component
extracts the user's intent from the transcribed text. This step
usually involves determining what action the agent should perform for the user, along with entities involved in that action (e.g., the action could be ``play,'' and the entity could be a song with title ``Shake It Off'').
NLU in the context of conversational AI is particularly challenging for several reasons. First, speech recognition
errors, as well as heterogeneous and informal styles of language use, often introduce noise to the user input and make understanding difficult. Secondly, many requests issued to digital assistants are brief and ambiguous, requiring an external knowledge source in order to select the most likely interpretation.
For example, a user query ``Play play that song train'' is difficult to comprehend because the sentence can be interpreted in several ways, especially if we consider the possibility of speech recognition errors and noisy user input. Is ``train'' the title of a song? Is ``play that song'' the song title and ``train'' the artist name\footnote{Ground truth: ``Play that song" is the name of a song by the band ``Train".}?
In order to play the correct song, the NLU component needs to correctly identify the entities and entity types in the request despite potential ambiguity, which is remarkably challenging.

In our work, we treat this parsing task as a sequence labeling problem performed by a bidirectional LSTM (BiLSTM) model \citep{DBLP:conf/icann/GravesFS05} (Section \ref{sec:multitaskModel}). This type of deep neural network based model requires a large amount
of training data to perform at high accuracy. 
As with many traditional machine learning problems, the granularity of the label space impacts the ease of the learning task as well as the cost of acquiring annotated labels. 
Coarse-grained labels are easier to obtain, e.g., via human annotation, and facilitate efficient model training. On the other hand, fine-grained labels are often more useful for downstream components in the AI system to consume in order to produce desired outcomes for the user. 
In the context of music entity labeling, an example of a \emph{coarse-grained} label is \emph{musicEntity}, which is a
collection of finer-granular music-related entities such as \emph{musicArtist}, \emph{musicAlbum}, and \emph{musicTitle}. In a coarse-grained label space, given the request ``Play the the kingdom of rain,''\footnote{Ground
  truth: ``Kingdom of Rain'' is the name of a song by the post-punk band ``The The''.}  the entire span of `the the
kingdom of rain'' will be labeled as one single \emph{musicEntity}. In contrast, in a \emph{fine-grained} label space,
``the the'' should be labeled as \emph{musicArtist} and ``kingdom of rain'' as \emph{musicTitle}. It is
apparent that fine-grained labels contain more detailed information about the true user intent and are more valuable for
the downstream components to take the accurate action. However,
correctly identifying the fine-grained entities in a user's request is time-consuming, costly, and often challenging
even for human annotators (e.g., requiring annotators to recognize idiosyncratic names of music artists), which leads to
insufficient hand-labeled training data for model training\footnote{An annotator may not know that ``The The'' is the name of a band and may provide incorrect fine-grained labels. As a result, it is often preferable for human annotators to annotate in a coarse-grained label space.}.

Our contribution in this paper is to describe a framework that leverages naturally occurring user behaviors to automatically annotate user requests with fine-grained entity labels. We use empirically validated heuristics to select user behaviors that indicate positive or negative engagement with content. These behaviors include tapping on content to engage with it further (positive response), listening to a song for a long duration (positive response), or interrupting content provided by the assistant and manually selecting different content (negative response). These user behaviors, which we refer to as \emph{user engagement signals}, provide strong indications of a user's true intent. We selectively harvested these signals
in a privacy-preserving manner to automatically produce ground truth annotations. Our solution only needs human annotators to provide coarse-grained labels, which are much simpler and faster to obtain with higher fidelity compared to a finer-grained labeling process. These simpler coarse-grained labels are then further refined using
user engagement signals, as explained in the following sections. Our framework is of particularly great value in scenarios where the conversational AI system extends to new domains or features, and corresponding training data need to be collected quickly and reliably for bootstrapping. Moreover, as will be illustrated shortly, user engagement signals can help us to identify where the digital assistant needs improvement by learning from its own mistakes. Our approach significantly increases the volume and quality of our training data without adding much annotation cost, nor jeopardizing user privacy or user experience.

In order to incorporate both coarse-grained labels (by human annotators) and fine-grained labels (inferred by our framework), we designed and deployed a \emph{multi-task learning} framework in our production environment, which treats coarse-grained and fine-grained entity labeling as two tasks. We also incorporated an external knowledge base consisting of entities and their relations to validate the model's predictions and ensure high precision. We show that our data generation framework coupled with these modeling and validation strategies leads to
significant accuracy improvements for both the coarse-grained and fine-grained labeling tasks. More importantly, we demonstrate that our framework yields significantly better user experience in a real-world production system.

\section{Related Work}

The use of unsupervised or weakly supervised data to improve performance in entity-labeling tasks 
has a long history. A well-established
strategy is to start with some seed examples and then use contextual features and co-training to identify and refine new
examples \citep{W99-0613,W14-1611}, building up a corpus that can then be used to train a model. In \citet{N15-1128}, the
authors show that distributed representations can further improve performance of such systems, and in \citet{C18-1196} this and two
related approaches are compared and found to outperform methods that do  not use distributed representations.

In recent work, \citet{yang-mitchell:2017:Long} describe an LSTM based architecture
that uses external resources like WordNet and a  knowledge base of triples {\it (entity1, relation, entity2)} to carry out entity labeling in two stages: first identifying chunks and second labeling them. By representing external
concepts via embeddings and training an attention mechanism, the system is able to leverage these concepts: the
attention mechanism serves partly to weight the appropriate sense of an ambiguous term, correctly distinguishing between
(for example)
`Clinton' as person or as a location depending on the context. Our use of a knowledge base is simpler than this,
essentially
acting as an existence check to re-rank alternatives produced by the model.

Improving performance of dialog systems by using information about user engagement and task completion is a standard technique for systems
that use reinforcement learning to acquire or improve a dialog policy: for a review see
\citep{DBLP:journals/pieee/YoungGTW13}, and for some recent developments \citep{DBLP:journals/csl/GasicMRSUVWY17}.
  However, to our knowledge, our work is the first to use inferences about task completion to derive training data for
  sequence labeling rather than policy learning.

\section{Generating Weakly Supervised Data}
\label{sec:dataAugmentation}
In this section, we describe user engagement signals as well as how we use them to generate fine-grained entity annotations. 
In the rest of the paper, we will use queries expressing a \emph{play music} intent as the example use case to illustrate our method \footnote{Note that in our setup, the NLU component contains a module that classifies requests into domains such as music. Although user engagement signals can be used to improve the domain chooser, this work focuses on improving the entity labeling component that follows.}. Our proposed methods can be extended straightforwardly to other domains where user engagement signals are available.

\subsection{User Engagement Signals}

User engagement signals refer to user behaviors that indicate whether the user feels positive or negative about the agent's chosen action, without the agent asking for explicit feedback.
In our scenario of the \emph{play music} intent, a positive signal is defined as the user listening to the song initiated by the agent for more than a threshold amount of time. We determined the threshold to be $30$ seconds by asking annotators to grade the success of a request and correlating the grades with how long a song was played (the vast majority of songs played $>30$ seconds were graded successful). A negative signal is defined as the user aborting the song and switching to a different one, or the user playing a desired song by searching for it manually after the agent claims it could not find the song\footnote{Although there are cases where a user changes her mind and aborts a correctly selected song, we find that the majority of cases where a user switches to a related song are genuinely unsuccessful cases.}.

\subsection{Engagement-Annotated Fine-Grained Data}

We first deploy a model based on human-labeled data in a coarse-grained label space.
This model infers a user's intent and passes it to the Action component. For example, given the request ``Play play that song train'', suppose the model predicts ``play that song train'' as \emph{musicEntity}, which is a coarse-grained label. Using our model,
we can obtain fine-grained labeled data in the following scenarios. 
The first scenario is that the downstream component makes a correct decision and plays the song ``Play that song'' by the artist ``Train''. If we receive a positive engagement signal from the user (i.e., this song was played for a certain amount of time), we can retrieve detailed metadata of the played song including the title, album, and artist. In this case, the title is ``Play that song'' and the artist is ``Train.'' We then map this fine-grained information back to the utterance to \emph{regenerate} fine-grained entity labels for each token. This results in a high quality training example that is automatically labeled with fine-grained entity types, where ``play that song'' maps to the \emph{musicTitle} type, and ``train'' maps to the \emph{musicArtist} type, in contrast to one single \emph{musicEntity} coarse label. 

\begin{figure}[h]
  \centering
  \includegraphics[width=1\textwidth]{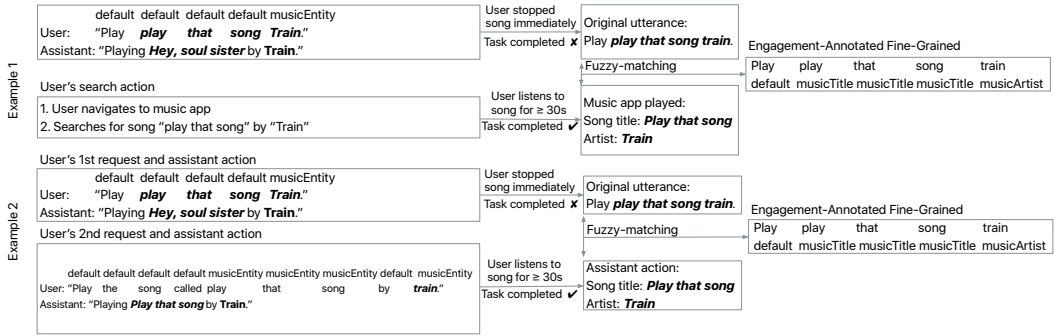}
  \caption{Examples of generating engagement-annotated fine-grained data.}
  \label{fig:autoNER}
\end{figure}

The second scenario is that the downstream component makes a wrong decision and returns undesired results, e.g., a song that the user does not want or misinterpreting the request to be for a movie instead. 
From our analysis, users will often immediately stop the incorrectly chosen content and manually search the intended song and then play it, or interrupt the content with a query that paraphrases the original query. This is a strong indicator that the NLU and downstream components failed to fulfill the user's intent, and the song manually played by the user (or played by the system following the paraphrase) is actually the desired one. Our model then utilizes metadata of the ultimately played song to gather the correct fine-grained entity labels. 

It is worth noting that the metadata of the song is standardized and contains properly spelled entity names, whereas the original utterance may be noisy and informal. In order to map the finer level music information back to the original utterance, we employ an edit-distance based \emph{fuzzy matching} algorithm to perform this mapping. The matched tokens are labeled as the identified entities if the fuzzy matching confidence score is above a threshold\footnote{We collected human judgments of similarity and found that strings with fuzzy matching confidence scores over $0.8$ tend to be rated as highly similar by humans. As a result, we used $0.8$ as the fuzzy matching threshold.} and the remaining tokens will be labeled as
  ``default" (i.e., meaning the token does not reference an entity). The fuzzy matching algorithm can tolerate spelling errors, missing or redundant tokens, and ordering problems, which frequently occur in conversational AI systems (e.g. matching ``your beautiful'' to ``you're beautiful,'' and ``this is you came for'' to ``this is what you came for'').
Figure~\ref{fig:autoNER} shows two examples of using the fuzzy matching algorithm to annotate an utterance that was originally predicted incorrectly. The error is then corrected by mapping the song title and artist name to the original utterance.

In summary, 
we describe two scenarios that provide us with valuable fine-grained entity labels: (1) queries with strong positive user engagement signals, and (2) queries with strong negative user engagement signals followed by the user's corrective action. Both cases will be leveraged by our model and framework to retrieve weakly-supervised and finer-granular ground-truth entity labels for the original user utterance. We refer to this fine-grained dataset enriched by user engagement signals as the \emph{engagement-annotated} data.
Since the \emph{engagement-annotated} data and \emph{human-annotated} data were labeled from different label spaces (fine-grained v.s. coarse-grained, respectively), it is not straightforward to incorporate these two training data sources together\footnote{We believe this is a realistic challenge in many scenarios: since fine-grained entity labeling is a more difficult and time-consuming task, it is easier to obtain high-quality human-annotated data with coarse-grained entity labels, whereas weak supervision may provide fine-grained (but potentially noisy) labels.}. In the following section, we introduce a multi-task learning approach that leverages both datasets jointly to improve entity recognition for both the coarse-grained and fine-grained labeling tasks.

\section{Multi-task learning}
\label{sec:multitaskModel}

\begin{figure}
  \centering
  \includegraphics[width=0.7\textwidth]{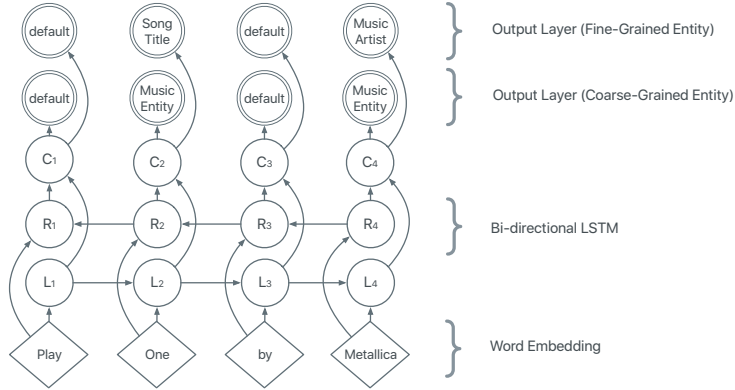}
  \caption{Main architecture of the multi-task learning network. Word and context feature embedding are given to a  bidirectional LSTM, where $L_i$ represents the word $i$ and its left context, $R_i$ represents the word $i$ and its right context. Concatenating these two vectors yields a representation of the word $i$ in its context $C_i$, which is fed into two independent output layers - one for coarse-grained entity typing task and the other for fine-grained entity typing task.}
  \label{fig:multiTask}
\end{figure}

We design a multi-task learning framework to better utilize engagement-annotated data (with finer-granular entity
labels) and human-annotated data (with coarse-granular entity labels). Note that
the same training example is not initially required to have both coarse-grained and fine-grained labels. 
As shown in Figure~\ref{fig:multiTask} , the multi-task learning model utilizes a deep neural network architecture based on bidirectional LSTMs (BiLSTM)
\citep{DBLP:conf/icann/GravesFS05}. For every query, we first generate a vector containing a list of customized
features representing domain and context information. These features are  pre-trained in an embedding layer for dimension reduction, such that each token in the utterance is represented by a word vector. The word embeddings are generated using word2vec \citep{DBLP:conf/nips/MikolovSCCD13} and are trained on data sampled from our production usage.  Both the reduced feature vector and the token word embeddings are passed to the BiLSTM as inputs for training. The outputs from the forward and backward pass of the first BiLSTM layer are concatenated to form the input for the second BiLSTM layer. This is followed by a linear projection layer and two softmax layers: one for predicting coarse-grained entity type labels and another for predicting fine-grained entity type labels. The loss function is defined as
{\small
\begin{equation}
\begin{aligned} 
\label{loss} 
L(\mathbf{w}) &= \mathds{1}_{\{d \in D_{\text{CG}}\}} \sum_i y_i \log p_i  + \mathds{1}_{\{d \in D_{\text{FG}}\}} \sum_j z_j \log q_j + \lambda \lVert \mathbf{w} \rVert^2 
\end{aligned}
\end{equation}
} where $d$ denotes sampled mini-batch; $D_{\text{CG}}$ denotes human-annotated coarse-grained data; $D_{\text{FG}}$ denotes engagement-annotated fine-grained data; $\mathbf{w}$ refers to network weights;  $\lambda$ denotes $L^2$ regularization parameter; $y$, $p$ refer to ground truth and predicted class for coarse-grained entity typing task; $z$, $q$ refer to ground truth and predicted class for fine-grained entity typing task.

For every iteration during training, we select a mini-batch ($d$) from one of the data sources based on a pre-defined sampling weight assigned to each source. If the mini-batch belongs to the human-annotated data ($D_{\text{CG}}$) which follows a coarse-grained entity label space, we perform a forward and backward pass through the input projection layers, LSTM network and the coarse-grained entity typing softmax output layer. If the mini-batch belongs to the engagement-annotated data ($D_{\text{FG}}$), we perform a forward and backward pass through the input projection layers, LSTM network and the fine-grained entity typing softmax output layer. Note that the lower level LSTM network is shared between both the tasks, and its weights are updated during every iteration. However, the weights of the coarse-grained entity typing and the fine-grained entity typing output layers are updated only when the mini-batch is sampled from the respective data source.
This multi-task framework effectively increases the training data size for LSTM layers and facilitates better feature representation to improve entity typing accuracy.

\section{Knowledge Base Validation}
\label{sec:kbValidator}

We can further improve our fine-grained entity labeling by utilizing an external knowledge base. For example, given the query ``Play something by the Beatles,'' we label ``something'' as \textit{musicTitle} partially because it exists as a song by \textit{The Beatles} in a music knowledge base. If the user had said ``Play something by Taylor Swift'', since artist \textit{Taylor Swift} has no song called ``something'', the system should interpret the utterance to mean ``play any song by the artist Taylor Swift'' instead. Therefore, an authoritative knowledge base containing relational information about music entities provides an efficient and robust way to validate our model.

During inference, after the model predicts the fine-grained entity label distribution for the sequence, we perform a beam search
over the prediction lattice and select the top five alternatives based on the average sequence-level probabilities. For
each alternative predicted by the fine-grained entity typing module, we construct a relational query and query the KB
validator to check whether it is a valid relational tuple. For example, if the model predicts  ``something'' as
\textit{musicTitle}, and ``the Beatles'' as \textit{musicArtist} for one of the candidate parses for ``Play something by the Beatles'', we construct a KB query
(``something",  ``the Beatles'') and query our KB validator. After performing this look-up for the top
sequence label alternatives, we send these as features to a hand-crafted re-ranking module which re-ranks the hypotheses and returns
the best alternative. For the utterance ``Play something by Taylor Swift'', if the model predicts ``something'' as
\textit{musicTitle} and ``Taylor Swift'' as \textit{musicArtist}, we would not find this tuple in the KB and would thus
down-rank this particular hypothesis. By using the KB validator, we are more likely to promote the correct fine-grained entity labels as the top prediction.

\section{Results}
\label{sec:result}

We use two separate test sets to evaluate the two tasks performed by the multi-task model: (i) \emph{coarse-grained
  blind test set} (ii) \emph{fine-grained blind test set}. Both test sets were constructed by randomly sampling from
production usage and then hand-annotated with ground truth labels. The coarse-grained test set was labeled using coarse-grained entity labels, and the fine-grained test set was labeled using fine-grained entity labels. 
From the coarse-grained test set, we compute the model's coarse-grained entity error rate (CGEER), which measures the utterance-level error rate on coarse-grained entity types. From the fine-grained test set, we compute the model's fine-grained entity error rate (FGEER), which measures the utterance-level error rate on fine-grained entity types. 

\subsection{Experiments}
For model training, we use mini-batch stochastic gradient descent with momentum and minimize the sum of the cross entropy loss across all the tokens in all the utterances. To reduce overfitting, we add $L^2$ regularization of \textbf{$\lambda$} = $0.0005$ and dropout \citep{JMLR:v15:srivastava14a}
of $0.25$. We begin model training with an initial learning rate of $0.9$ and employ a scheduled learning rate decay approach with a factor of $0.8$ at the end of every epoch. 
To assign the sampling weights for human-annotated training data and engagement-annotated data, we perform a grid search with a step size of 0.1, with both weights ranging between (0, 1) and summing to 1. We got the best results with sampling weight of $0.5$ for human-annotated and $0.5$ for engagement-annotated data (i.e. equal probability of selecting a mini-batch belonging to one of the data sources). All other hyperparameters were also obtained after extensive grid search\footnote{We varied the initial learning between [0.5, 1] with a step size of 0.1, learning rate decay factor between [0.5, 1] with a step size of 0.1 and dropout between (0, 0.5) with a step size of 0.05.} on a held out development set for lowest utterance error rate.

To understand the impact of the human-annotated and engagement-annotated data on coarse and fine-grained entity labeling, we compare our model accuracies across different training data settings. We vary the amount of human-annotated data -- $5k$, $10k$, $30k$, $60k$, $90k$ and $120k$ -- and engagement-annotated data -- $0$, $260k$, and $520k$ -- and compare the CGEER and FGEER in the various settings. Note that in the baseline settings (where no engagement-annotated data is added), we only report CGEER, because the baseline model is trained purely on human-annotated coarse-grained data and hence can predict only coarse-grained entities. 

\subsection{Discussion}

\begin{table}
\small
\captionsetup{font=footnotesize}
  \caption{{\it Impact of adding 1x (260k training examples) and 2x (520k training examples) Engagement-Annotated Fine-Grained data on Coarse-Grained Entity Error Rate (CGEER) and Fine-Grained Entity Error Rate (FGEER). Note that since the baseline models are trained only on human-annotated coarse-grained data and cannot predict fine-grained entities, we only report CGEER. We report the mean $\pm$ standard error of the mean (SEM) across $21$ runs. All CGEER improvements over the baseline are significant ($p < 0.005$).}}
  \label{geer}
  \centering
  \footnotesize
  \begin{tabular}{llrrr}
  
    \toprule
    \multicolumn{5}{c}{Coarse-Grained and Fine-Grained Entity Error Rates}                   \\
    \cmidrule(r){1-5}
    Human- & Engagement- & CGEER(\%) & FGEER (\%) & FGEER (\%)\\
    \vspace{0.1mm}
    Annotated & Annotated & & & (+ KB Validator) \\
    \vspace{0.1mm}
    Coarse-Grained & Fine-Grained & & &  \\
    \midrule
    5k \hspace{2.5mm} (baseline)& & 10.71 $\pm$ 0.05 & -- & --\\
   5k & + 1x & 9.58 $\pm$ 0.03 & 17.95 $\pm$ 0.07 & 17.47 $\pm$  0.12 \\
   5k & + 2x & \textbf{9.51} $\pm$ 0.04 & \textbf{17.40} $\pm$ 0.07 & \textbf{17.24} $\pm$ 0.16\\
    \midrule
     10k \hspace{1mm} (baseline) & & 9.89 $\pm$ 0.04 & -- & -- \\
    10k & + 1x & \textbf{9.04} $\pm$ 0.05 & 17.89 $\pm$ 0.07 & 17.42 $\pm$ 0.02 \\
    10k & + 2x & 9.09 $\pm$ 0.06 & \textbf{17.54} $\pm$ 0.07 & \textbf{17.05} $\pm$ 0.13\\
    \midrule
     30k \hspace{1mm} (baseline)& & 8.69 $\pm$ 0.03 & -- & -- \\
     30k & + 1x & \textbf{8.25} $\pm$ 0.03 & 17.82 $\pm$ 0.09 & 17.26 $\pm$ 0.11\\
    30k & + 2x & 8.29 $\pm$ 0.04 & \textbf{17.10} $\pm$ 0.07 & \textbf{16.79} $\pm$ 0.11\\
    \midrule
    60k \hspace{1mm} (baseline)& & 8.05 $\pm$ 0.04 & -- & -- \\
    60k & + 1x & \textbf{7.79} $\pm$ 0.04 & 17.92 $\pm$ 0.09 & 17.16 $\pm$ 0.12\\
    60k & + 2x & 7.87 $\pm$ 0.04 & \textbf{17.01} $\pm$ 0.10 & \textbf{16.83} $\pm$ 0.07\\
    \midrule
    90k \hspace{1mm} (baseline)& & 7.72 $\pm$ 0.03 & -- & -- \\
    90k & + 1x & \textbf{7.48} $\pm$ 0.03 & 17.73 $\pm$ 0.07 & 17.42 $\pm$ 0.08\\
    90k & + 2x & 7.50 $\pm$ 0.04 & \textbf{17.46} $\pm$ 0.08 & \textbf{16.66} $\pm$ 0.13 \\
    \midrule
    120k \hspace{0.1mm}(baseline)& & 7.63 $\pm$ 0.04 & -- & -- \\
    120k & + 1x & \textbf{7.46} $\pm$ 0.03 & 17.86 $\pm$ 0.06 & 17.06 $\pm$ 0.12\\
    120k & + 2x & 7.56 $\pm$ 0.04 & \textbf{17.62} $\pm$ 0.10 & \textbf{16.69} $\pm$ 0.24\\
    \bottomrule
  \end{tabular}
\end{table}

\begin{table}

\captionsetup{font=footnotesize}
  \caption{{\it Comparison of Fine-Grained Entity Error Rates (FGEER) with and without KB Validator, where at least one of the top model predictions passes the KB validator. A model prediction can pass KB validator only if it contains two or more predicted entities (e.g. "Play \emph{X} by \emph{Y}", which allows  KB to validate the relation between the entities \emph{X} and \emph{Y}). Note that since KB Validator only activates for fine-grained entity types, we do not report results for the baseline model, which can predict only coarse-grained entities. We report the mean $\pm$ standard error of the mean (SEM) across $5$ runs.}}
  \label{kbtable}
  \centering
  \footnotesize
  \begin{tabular}{llllr}
    \toprule
    \multicolumn{5}{c}{Fine-Grained Entity Error Rates for KB Validator activated cases (\textasciitilde16\% coverage)}                   \\
    \cmidrule(r){1-5}
    Human-Annotated & Engagement-Annotated & KB Validator & FGEER (\%) & FGEER (\%)\\
    \vspace{0.1mm}
    Coarse-Grained & Fine-Grained & Activation (\%) & & (+ KB Validator)\\
    \midrule
    5k & + 1x & 16.03 & 6.38 $\pm$ 0.33& \textbf{3.28} $\pm$ 0.09\\
    5k & + 2x & 16.00 & 5.91 $\pm$ 0.24& \textbf{3.22} $\pm$ 0.17\\
    \midrule
    10k & + 1x & 16.06 & 6.54 $\pm$ 0.36& \textbf{3.23} $\pm$ 0.12\\
    10k & + 2x & 16.00 & 6.59 $\pm$ 0.43& \textbf{3.04} $\pm$ 0.07\\
    \midrule
    30k & + 1x & 16.02  & 6.67 $\pm$ 0.19 & \textbf{3.13} $\pm$ 0.04 \\
    30k & + 2x & 16.07 & 5.52 $\pm$ 0.20& \textbf{3.07} $\pm$ 0.12 \\
    \midrule
    60k & + 1x & 16.08 & 6.64 $\pm$ 0.56 & \textbf{3.24} $\pm$ 0.11\\
    60k & + 2x & 16.03 & 6.19 $\pm$ 0.37& \textbf{3.21} $\pm$ 0.13\\
    \midrule
    90k & + 1x & 16.04 & 6.53 $\pm$ 0.13& \textbf{3.38} $\pm$ 0.15\\
    90k & + 2x & 16.06 & 6.59 $\pm$ 0.27& \textbf{3.19} $\pm$ 0.13\\
    \midrule
    120k & + 1x & 16.04 & 6.85 $\pm$ 0.23 & \textbf{3.38} $\pm$ 0.10\\
    120k & + 2x & 15.71 & 6.87 $\pm$ 0.41& \textbf{3.33} $\pm$ 0.11\\
    \bottomrule
  \end{tabular}
\end{table}

From Table~\ref{geer}, we observe that adding $1\times$ engagement-annotated fine-grained data ($260k$ training examples) consistently reduces CGEER compared to the baseline for all amounts of human-annotated data (significant across $21$ model runs; $p < 0.005$). In addition, adding weakly supervised fine-grained data has a larger impact when there is a relatively small amount ($5k$) of human-annotated data (\textasciitilde1.1\% absolute error reduction in CGEER) compared to when there is a large amount ($120k$) of human-annotated data (\textasciitilde0.2\% absolute error rate reduction). We notice that adding $2\times$ engagement-annotated data does not reduce CGEER more than the $1\times$ setting. We hypothesize that the reason for this could be that coarse-grained entity typing is an easier task, and there may be an upper-bound of benefits from adding engagement-annotated data. On the other hand, we observe that adding $2\times$ engagement-annotated data helps improve generalization for the fine-grained entity typing task and consistently reduces FGEER compared to the $1\times$ setting ($p < 0.005$). 

Using the KB validator to re-rank the fine-grained predictions further reduces FGEER in all training data settings (Table~\ref{geer}). Table~\ref{kbtable} shows results from additional experiments where we specifically investigated the effect of the KB validator. Among examples where any of the top model hypotheses passed the KB validator by containing two or more related entities (\textasciitilde16\% of all examples in the test set), FGEER dropped from between $5.52\%-6.87\%$ to between $3.04\% - 3.38\%$, a \textasciitilde50\% relative error rate reduction. These results suggest that the KB validator successfully incorporates known relationships between entities to identify correct labels from top  model hypotheses.

An important observation from these results is that adding engagement-annotated data achieves a similar effect on the original coarse-grained prediction task as adding more human-annotated training data. As observed in Table~\ref{geer}, adding $1\times$ engagement-annotated data to the $5k$ human-annotated gives us more accuracy gains 
than adding an additional $5k$ human-annotated training data, and adding $1\times$ engagement-annotated data to the $90k$ human-annotated data gives us more accuracy gains than an additional $30k$ human-annotated data.
This comparison suggests that we can improve entity labeling accuracy at lower costs by leveraging automatically-obtained user engagement signals. These implications are particularly promising for low-data settings (such as when new features or domains are launched), where hand-labeling training data can be expensive and time-consuming.

One of our motivations for enabling the system to produce fine-grained entity labels is the expectation that a more granular representation of the user's intent would increase the likelihood of the downstream component selecting the correct action and lead to better user-facing results. 
To test whether this expectation holds true, 
we sampled approximately $5k$ utterances from production usage that express a ``play music'' intent and contain references to multiple entities and sent them hrough the full system end-to-end. All system components were held constant except the NLU module, which either contained the baseline model (which produced coarse-grained labels) or the enhanced model with KB validation (which had the ability to produce and validate fine-grained labels). We asked annotators to grade the response returned by the system as ``satisfactory'' or ``unsatisfactory'' given the user request and computed the percentage of "unsatisfactory" grades given the baseline and enhanced models, which we call \emph{task error rate}.
Results produced by the enhanced model achieved a relative task error rate reduction of $24.64\%$. This suggests that using engagement-annotated data combined with a knowledge base validator produces high-quality fine-grained predictions that are easier for downstream components to consume, ultimately leading to a better user experience.

We observe that our model improves user-facing results especially for requests that contain difficult or unusual language patterns. For example, the enhanced system correctly handles queries such as ``Can you play Malibu from Miley Cyrus new
album'' and ``Play Humble through my music Kendrick Lamar''. Also, the enhanced model identifies entities that users are more likely to refer to in cases of genuine linguistic ambiguity. For example, in ``Play one by Metallica'', ``one'' could either be a non-entity token (meaning play any song by Metallica), or it refer specifically to the song called ``One'' by ``Metallica''. Since most users listen to the song ``One'' by the ``Metallica'' whenever they
say ``Play one by Metallica'', our model trained on engagement-annotated data will learn to predict ``one'' as \textit{musicTitle}, thus better capturing trends and preferences in our user population\footnote{There may be cases where an individual user's preference is overwritten by the population preference. For example, a user may actually want \emph{any} song by Metallica when they say ``Play one by Metallica''. Future work could explore using individual users' engagement behaviors to better address personalization.}.

\section{Conclusions}

In this paper, we described a system that leverages user engagement signals to produce fine-grained entity labels for a sequence labeling task in the context of a conversational AI agent. We showed that combining this engagement-annotated data with human-annotated coarse-grained data in a multi-task framework significantly improves accuracy for both coarse-grained and fine-grained entity labeling tasks. We also showed that using a knowledge base to validate and select entity labels from the model's top predictions brings additional accuracy gains. We further tested the impact of our changes on the system end-to-end and showed that they significantly improved user-facing results. We believe our method can be applied in many practical settings to speed up new feature deployment (especially when little human-annotated data exists), and ultimately improve user experience.

\small

\microtypesetup{protrusion=false}
\bibliography{Biblio}
\bibliographystyle{acl_natbib}

\end{document}